# Ultrasound Elastography: Excitation Methods and Low-Cost Alternative

Corin F. Otesteanu
ETH Zurich, Switzerland
*corino@ethz.ch*

***Abstract*** **– Ultrasound elastography (or elasticity imaging) is a method that characterizes the tissue mechanical properties with the purpose of aiding the diagnosis of certain pathologies. Many novel clinical applications use the assumption that these tissue properties are changed in the presence of pathological processes. Characterization of tissue mechanical properties is achieved by imaging the response of the tissue to an induced stress or excitation. A plethora of approaches have been used in the scientific community to induce excitation in tissue, which usually require additional tools or devices than are costly. Furthermore these are usually integrated in the imaging pipeline using additional software or hardware for control. An alternative that is simpler, faster to integrate (and low-cost) is proposed in this work, by using readily available smartphones as a source of mechanical excitation. The induced displacements are used to solve the elasticity inverse problem (IP) using a finite element method (FEM) approach. Preliminary elasticity reconstructions of a phantom that mimics tissue are similar to when using a dedicated actuator as means of inducing harmonic displacements, illustrating the feasibility of our low-cost ultrasound elasography approach.**

## I. Introduction

Elastography (elasticity imaging) [1], is a family of approaches which allow the measurement of tissue mechanical properties in a noninvasive manner (e.g. shear modulus, Young's modulus) for the purpose of aiding diagnostic procedures of pathological processes [2].

The elastography procedures consists in three steps. First, a perturbation of the tissue is induced by applying a quasi-static [3], harmonic [4], [5] or transient [6], [7] stress. Then, the resulting mechanical response can be measured by imaging techniques like ultrasound [1], or magnetic resonance imaging [2]. Finally the biomechanical properties of the underlying tissue are estimated by using a mathematical model based on the wave equation of motion and on a constitutive equation relating stress and strain [8].

A plethora of methods have been used throughout the literature to induce displacements inside tissue, when using ultrasound elastography. Most commonly, these were produced using an acoustic or mechanical (see Fig. 1) source of excitation. A brief overview of these methods is given below, for a more in depth comparison of elastography methods, see the review articles [1], [2]. Acoustic radiation force methods rely on the momentum transfer between the acoustic wave and a target location inside the tissue, creating a transient displacement. Acoustic Radiation Force Impulse (ARFI) imaging is a method based on radiation force, where short duration radiation forces (called pushes) created by an ultrasound transducer are used to generate local displacements in tissue [7]. The same transducer is used in imaging mode to track the displacements. A 2D region of interest (ROI) is mapped by successive repetitions of the previously mentioned cycle, yielding qualitative [9] (ARFI) or quantitative (SWEI) (after further processing) elastograms [10]. Supersonic shear imaging (SSI) is also an approach based on radiation force, where several shear-waves interfere constructively to create cylindrical shear waves, allowing for a larger area to be imaged. The wave propagations are imaged by an ultrafast scanned (plane wave imaging) and quantitative elasticity maps are deduced from their speed of propagation [6].

In contrast to acoustic radiation force based elastography methods, where the source of excitation are the acoustic waves created by the transducer, mechanical based methods rely on external sources of vibration to produce the desired displacements. Many variations have been implemented throughout the literature. FibroScan is a commercially available ultrasound elastography machine developed by Echosens with a single element transducer that has an integrated mechanical exciter, where a low-frequency (50 Hz) mechanically generated impulse is used to induce a shear-wave in tissue. Quantitative 1D elasticity maps are then deduced from the shear-wave speed [11]. In [4], in order to generate harmonic shear waves, a custom made patient bed with an integrated, software controlled, loudspeaker was used to couple vibrations into the tissue. In [12] a custom built shaker, consisting of two voice coil motors controlled by a driving circuit that could be attached to the probe to generate harmonic excitations was used. Other groups used commercially available mechanical vibration devices controlled by a signal generator or microcontroller to generate shear waves [13] (single shaker) or crawling waves [14] (two shakers standing on opposing sides of ultrasound transducer) in tissue.

In this preliminary study, we aim at developing a low-cost, and easily to integrate approach to induce harmonic excitations in ultrasound elastography, with devices that are readily available and do not need additional software or hardware integration with the imaging pipeline.

## II. Methods

Smartphones, devices that are nowadays commonly available, could be used to elicit harmonic excitations using vibrations in a frequency range of approximately 100 - 200 Hz [15], depending on the smartphone. With this purpose, a Samsung Galaxy S4mini device was fixed to our ultrasound transducer using a velcro-band to provide harmonic excitation. The experimental setup can be seen in Fig. 2. A precise alignment between probe and phone was not required.

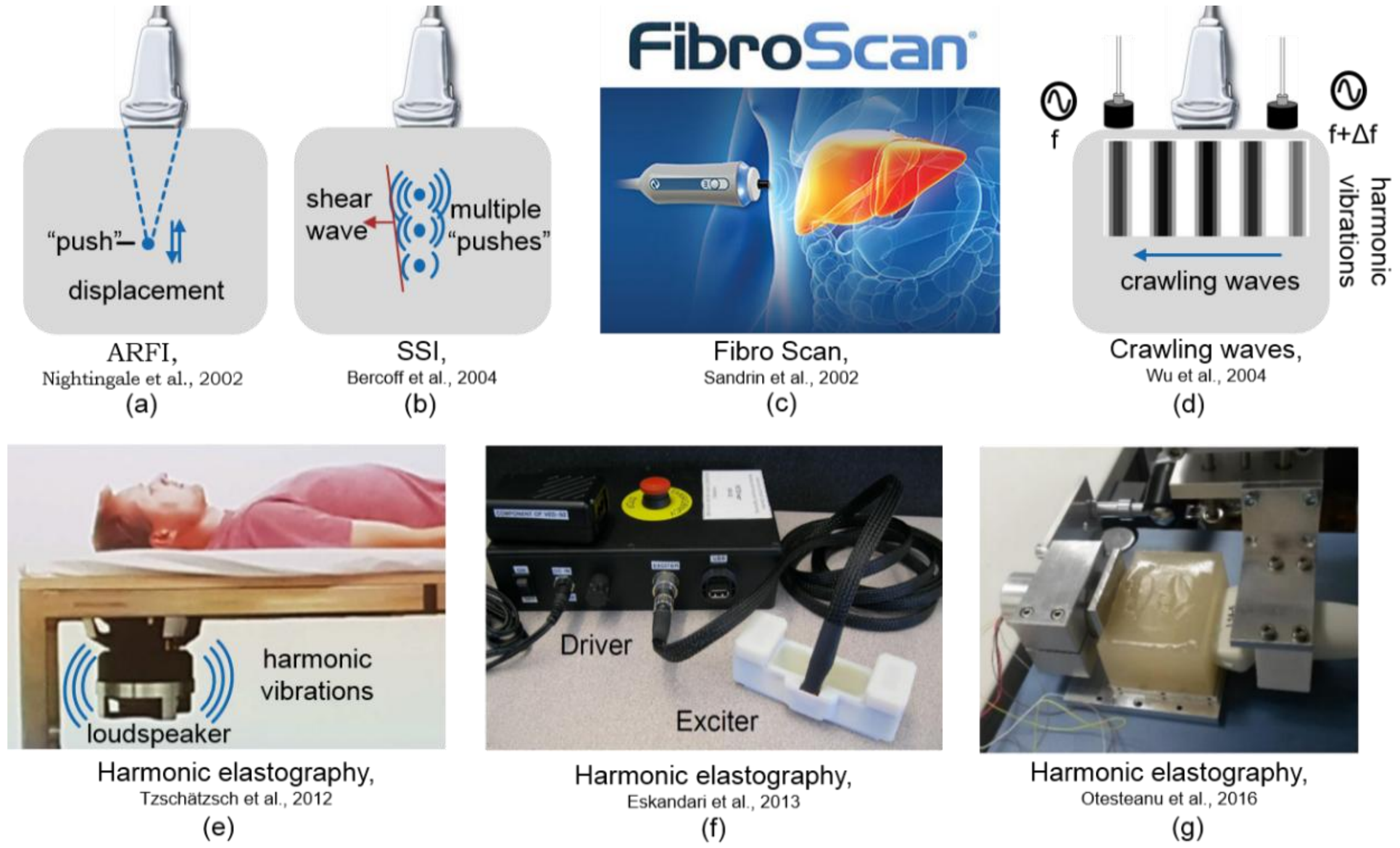

Figure 1. Excitation methods in ultrasound elastography achieved by acoustic force radiation: (a) Acoustic Radiation Force Impulse (ARFI) Imaging [9], (b) Supersonic shearwave imaging [6]; by transient mechanical impulse (c) FibroScan [11]; by two external harmonic exciters - crawling waves (d) [14], by single external harmonic exciters: (e) [4], (f) [12], (g) [13]. Figure 1(f) and (g) were reproduced with permission from IEEE, ©2014 IEEE [12] and ©2016 IEEE [13], respectively.

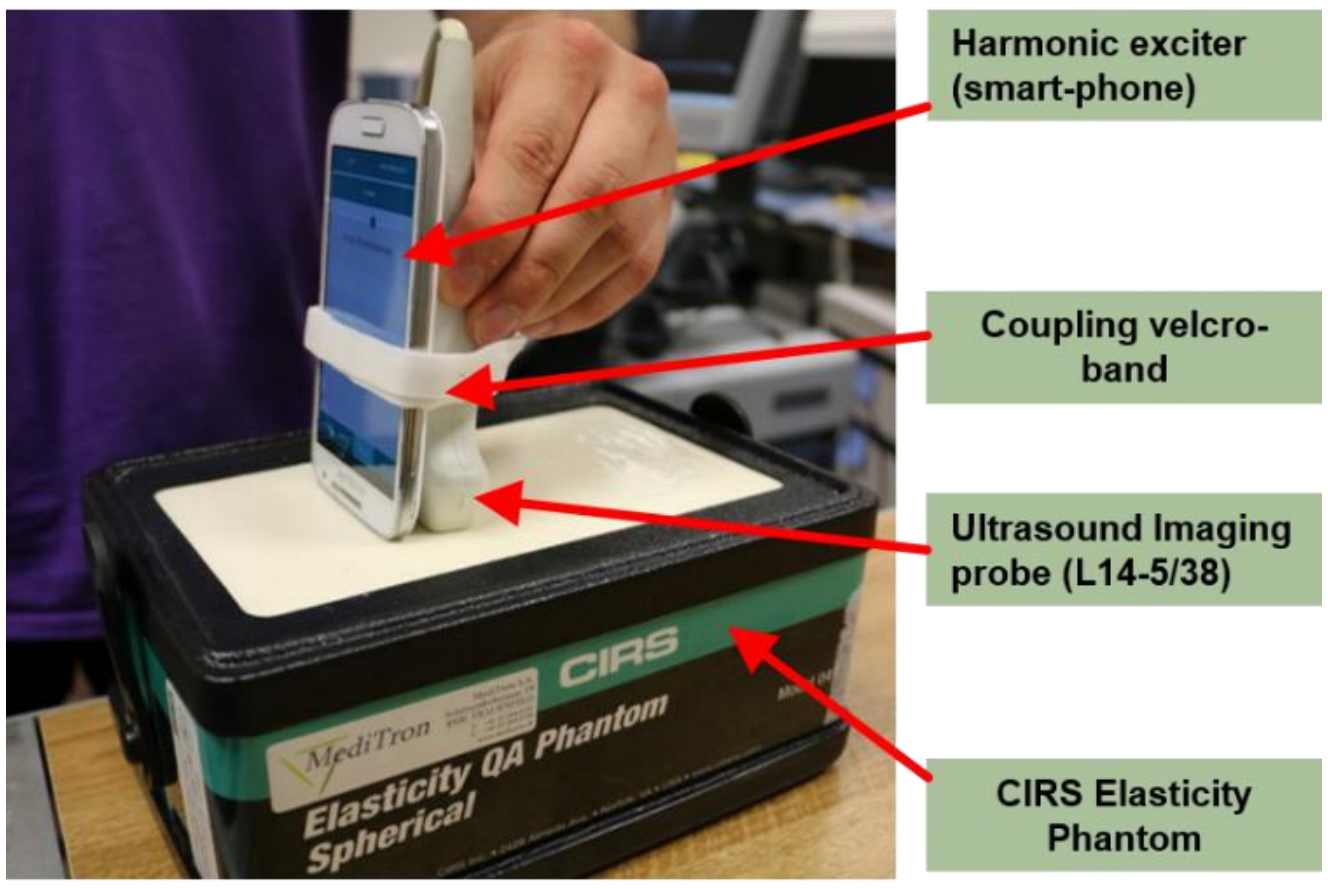

Figure 2. Measurement setup with a smartphone connected to an ultrasound probe via a velcro-band.

For proof-of-concept purposes, ultrasound elastography was performed on a CIRS quality assurance elasticity phantom, having a 1 cm diameter inclusion with a stiffness (Young's modulus) of 80 kPa, located in a homogenous 23 kPa background at a depth of 1 cm, according to the calibration sheet provided by the manufacturer. A SonixTouch Ultrasound machine (BK Medical Inc.) was used together with a SonixDAQ receiver module that allowed for high frame rate ultrasound images to be acquired. A flowchart of the acquisition and post-processing steps can be seen in Fig. 3. A L14-5/38 128-element linear-array ultrasound probe having the center frequency of 5 MHz with plane-wave transmits was used to image with a 1 kHz frame rate displacements induced by the smartphone. The induced displacements were tracked axially form the RF (radio-frequency) echo data with a time-domain cross correlation approach [16] (see Fig. 4). Displacement phasors were calculated using the Fourier transform of the displacement (wave propagation) data. After filtering around the frequency of interest (vibration frequency) they were used for elastic parameter reconstruction. For when the frequency of interest is not known (as can be the case for a smartphone) a peak detection algorithm from the MATLAB signal processing toolbox was used to identify it based on relative amplitude, inside a frequency range at which smartphones are known to vibrate. For reconstruction purposes, the equation of motion for a linear elastic and isotropic motion (1) is commonly used [2], [4], [12], [18].

$$\nabla \cdot [\mu(\nabla \boldsymbol{u} + (\nabla \boldsymbol{u})^T) + \lambda(\nabla \cdot \boldsymbol{u})I] = \rho \ddot{\boldsymbol{u}} \quad (1)$$

Here μ denotes the tissue shear modulus, **u** denotes the vector of displacement, **ü** the second time derivative of the displacement, λ denotes the first Lamè parameter and ρ is material density.

Herein a discretized, finite element method was used (2), where the Young's modulus was reconstructed in an inverse problem (IP) approach, as described in [17], [18].

$$[\boldsymbol{K}(\boldsymbol{p}) - \omega^2 \boldsymbol{M}]\boldsymbol{u} = \boldsymbol{f} \quad (2)$$

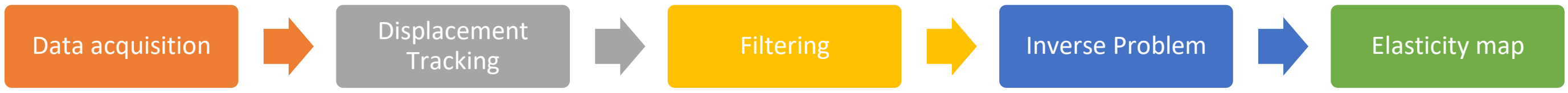

Figure 3. Flowchart describing the harmonic elastography reconstruction process: (i) data acquisition, (ii) displacement tracking of RF (echo) data, (iii) filtering of diplacment data at the excitation frequency (freqeuncy of interest) to obtain displacement phasors, (iv) using the displacement pahsors in an inverse problem approach to obtain (v) elasticity maps.

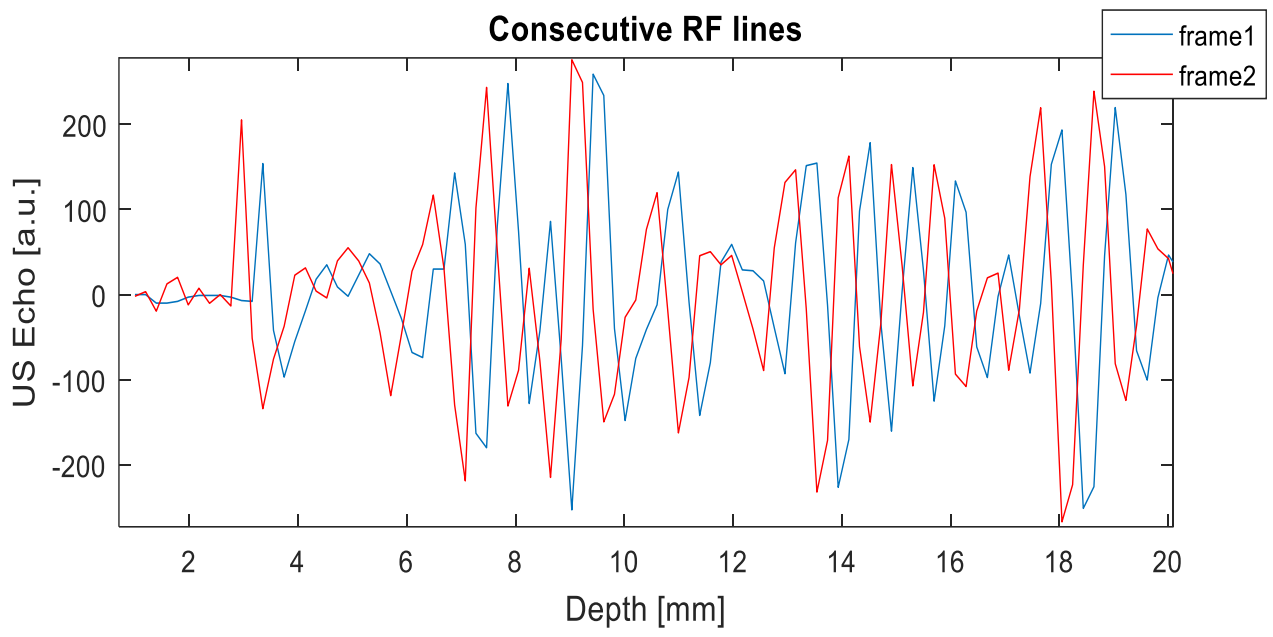

Figure 4. Consecutive ultrasound frames (lines) used for displacement estimation.

with **K** denoting the stiffness matix which dependents on the elastic parameters **p** (such as Young's modulus and Poisson's ratio), **M** denotes the (constant) mass matrix, ω is the excitation frequency, **u** denotes the displacement phasors, and **f** being the forces interacting with the object.

The IP can be formulated as an optimization where the goal is to find the elastic parameteres (shear modulus) that minimizes the difference between the model displacements $\hat{\mathbf{u}}(\mathbf{p})$ and the observed (measured) displacements $\hat{\mathbf{d}}$ **(p)** in tissue:

$$\mathbf{p} = \underset{p \geq 0}{\operatorname{argmin}} ||\hat{\boldsymbol{u}}(\mathbf{p}) - \hat{\boldsymbol{d}}||_2^2 + \alpha ||\boldsymbol{\Delta}\mathbf{p}||_2^2 \quad (3)$$

This discrepancy between measured and model displacements are allowed to account for potential noise present in the displacement. To improve the conditioning of the problem, L2 norm regularization is used, here **Δ** denoting the discrete Laplacian operator and α, a regularization coefficient. For every iteration in the optimization, the FEM model displacement $\hat{\mathbf{u}}(\mathbf{p})$ was updated using the current estimate of the elastic parameters in equation (2). To solvethe IP defined in (3) the trust-region reflective algorithm from the MATLAB optimization toolbox was used in this work.

To quantitatively asses the quality of the reconstruction, the following error-metrics were used: contrast-to-noise ratio (CNR), for characterizing image (inclusion) sharpness, and root-mean-square-error (RMSE), for characterizing deviations from ground-truth values, defined below:

$$\mathrm{RMSE} = \frac{||\boldsymbol{\mu} - \boldsymbol{\mu}_T||_2}{\sqrt{n}} \quad (4)$$

$$\mathrm{CNR} = \frac{2(\mu_{\mathrm{inc}} - \mu_{\mathrm{bg}})^2}{\sigma_{\mathrm{inc}}^2 + \sigma_{\mathrm{bg}}^2} \quad (5)$$

with **μ** denoting the shear modulus, $\boldsymbol{\mu}_T$ is the ground-truth shear modulus, $\mu_{inc}$ and $\mu_{bg}$ are the average stiffness values for the inclusion and background areas, and $\sigma_{inc}$ and $\sigma_{bg}$ denote the corresponding standard deviations.

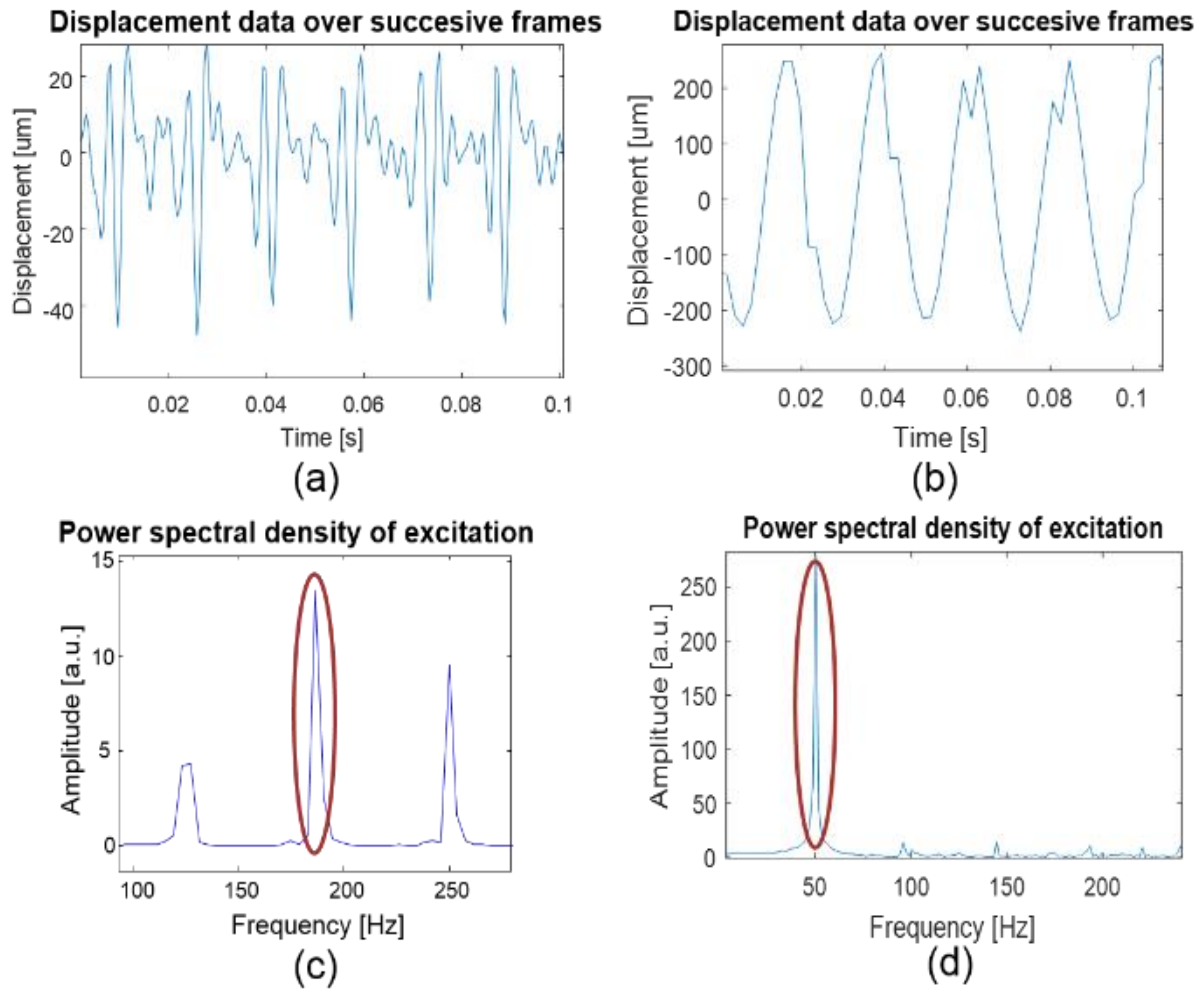

Figure 5. Harmonic displacements induced by smartphone (a) and dedicated actuator with position control (b) and their corresponding power spectral densities (c) and (d).

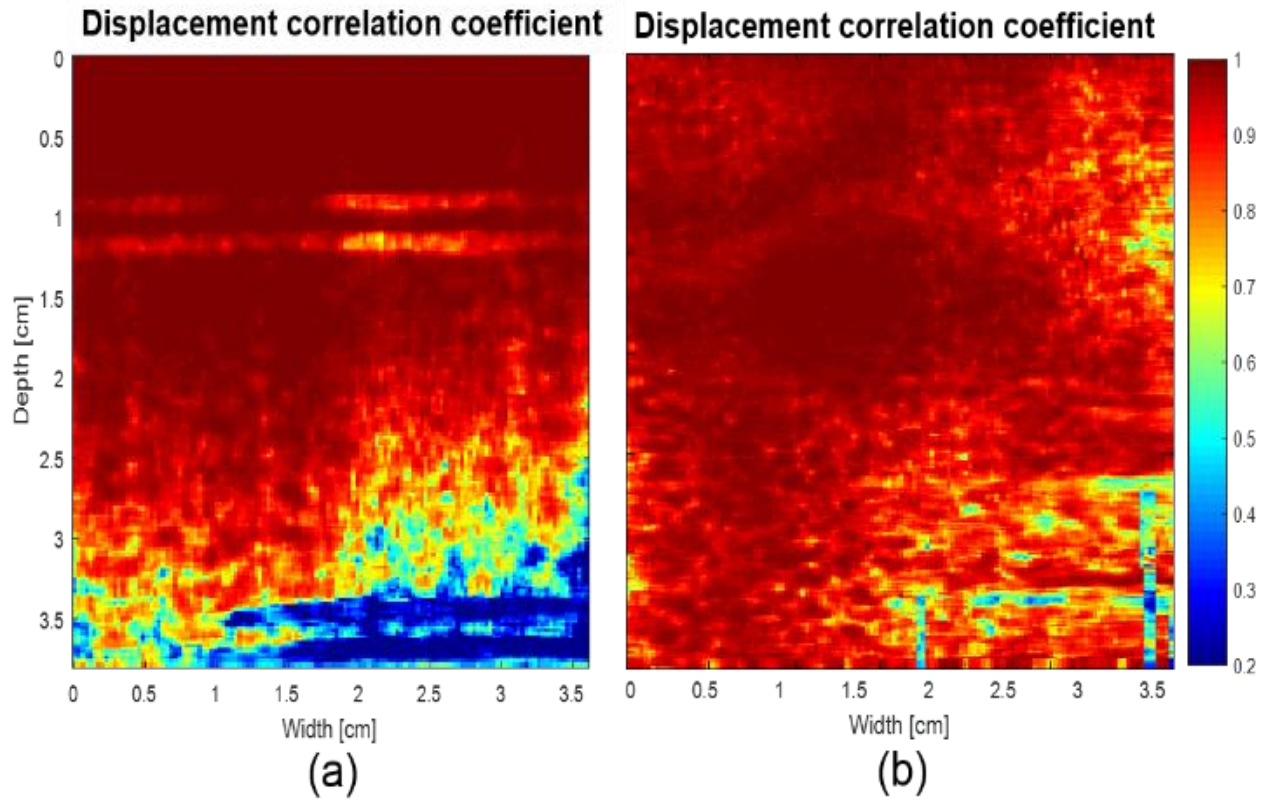

Figure 6. Cross-correlation coefficient maps calculated on the harmonic displacements induced by smartphone (a) and dedicated actuator with position control (b).

## III. Results and Discussion

The displacements induced by the smartphone and imaged with the ultrasound probe can be seen in Fig. 5 (a). Due to the nature of vibrations, displacements are small (~1-10 mm) and while the excitation produced is harmonic, it does not necessarily have a single sinusoidal component. Because of this, Fig. 5 (c) shows the power spectral density of the mobile phone, where several harmonic or noisy peaks can be observed in addition to the one corresponding to the excitation freqeuncy. For comparison purposes, the displacements and power spectral density of displacements induced by a dedicated voice coil actuator with position control fixed for a 50 Hz excitation are presented in Fig. 5 (b) and (d), respectively. An order of magnitude difference can be observed when comparing the two elastography excitation methods, which is to be expected due to the smaller vibration amplitude of the smartphone. Fig. 5 (a) and (b) also shows that while displacements in both cases are periodic, the one induced by the dedicated actuator is almost purely sinusoidal, as opposed to a sum of many noisy harmonics in the smartphone case.

TABLE I. RECONSTRUCTED CIRS PHANTOM PARAMETERS VALUES OBTAINED USING A SMARTPHONE AND DEDICATED ACTUATOR AND THE MANUFACTURER STATED VALUES (GROUND TRUTH).

| | Measurement Smartphone | Measurement Actuator | Ground Truth |
|---|---|---|---|
| Inclusion [kPa] | 91 | 84 | 80 |
| Background [kPa] | 15 | 19 | 23 |
| CNR [dB] | 24.9 | 22.8 | - |
| RMSE [kPa] | 14 | 13.2 | - |

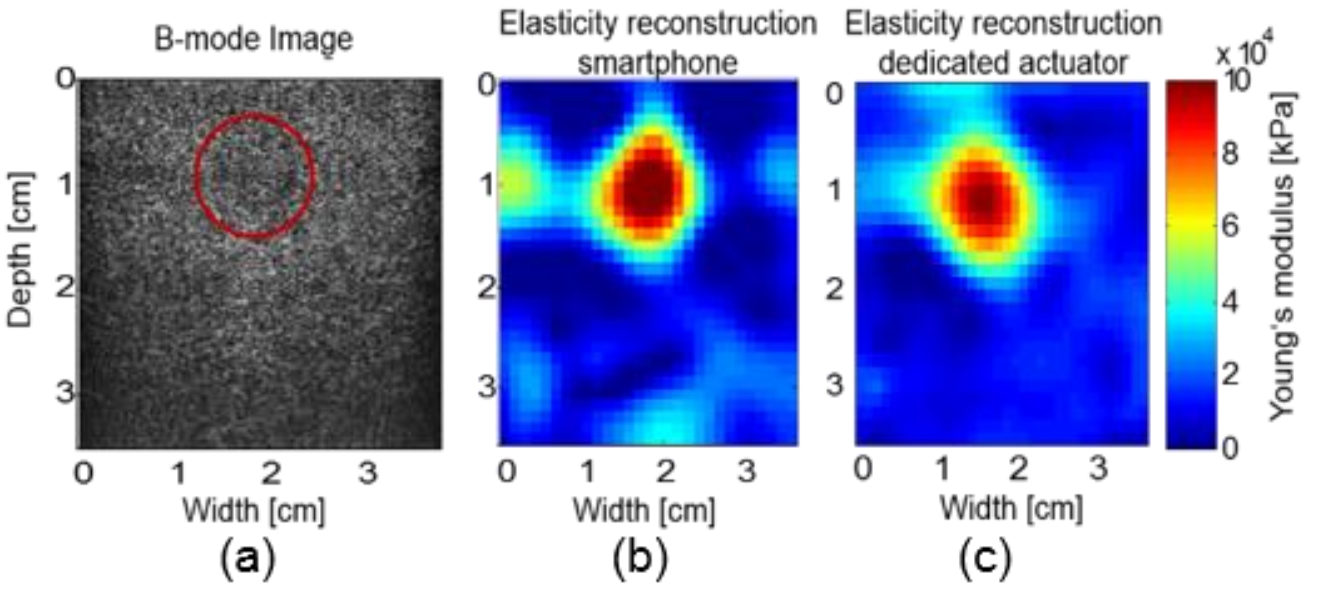

Figure 7. CIRS elasticity phantom with 1cm diameter circular inclusion: (a) inclusion marked on B-mode image and reconstructed elasticity map using smartphone (b) and dedicated actuator (c) as means of excitation.

Nevertheless, even though the displacements are significantly smaller in the smartphone case, they were still tracked accurately (with a correlation factor of >80%) up to a depth of 3.5 – 4cm. Fig. 6 displays the correlation factors for the induced displacements in the smartphone and dedicated actuator case. It can be seen that in the case of the dedicated actuator, displacements were tracked with a higher accuracy, especially in the bottom region of the figure (between 3 and 4 cm), with correlation values of 51% for the smartphone and 83% for the dedicated actuator. This can be explained by a combination of factors: on one hand the small vibration amplitude of the smartphone, and on the other the higher frequency of vibrations in the smartphone case, which are known to dampen faster with an increase in frequency, resulting in an overall lower SNR. Nevertheless, the proposed appraoch could be applied to surface organs, such as skin or breast. While a dedicated actuaor could be more versatile, using a smartphone, or any electronic device with a vibration function (mobile phone, pager etc) has the advantage of immediate applicability, without the need for hardware control using external devices (power supplies, amplifiers, microcontorllers) nor the need for complex custom built software applications for control. Moreover, no additional equipment would be needed for a patient examination using the proposed approach, compared to when using the dedicated actuator. Therefore, the low-cost term used in the paper title does not only refer to the price of the devices used, but also to the fact they easily available and ready to be used without additional time or finanacial overhaid due to hardware and software integration and control challenges. It is worth mentioning that for both appraoches, an ultrasound device with high frame rate capabilties that gives access to raw RF (echo) data is required, such as the devices often used in research or clinical settings with a focus on research (offered by BK Medical, Verasonics, Supersonic Imagine, etc).

For elasticity reconstruction, the displacements corresponding to the highest frequency peak (at 180 Hz) were filtered and used in the IP optimization problem (3). Fig. 7 shows the B-mode (ultrasound acoustic impedance) image of the CIRS elasticity phantom where the inclusion is barely visible and the corresponding reconstructed elasticity image. The circular inclusion was marked on the B-mode image with a red disc. In contrast to the B-mode image, the inclusion could be clearly identified in the elasticity map, with its position, shape and size corresponding to the ground-truth. The reconstructed mean values of Young's modulus for the inclusion and background were 91 kPa and 15 kPa, respectively, comparable to the manufacturer stated values.

Using the metrics defined by (4) and (5), an RMSE of 14 kPa and a CNR of 24.8 dB and was obtained for the reconstructed image, similar to those obtained when using a dedicated actuator, with an RMSE of 13.2 kPa and a CNR of 22.8 dB. Note that in the previously mentioned case the vibration frequency of the phone was 180 Hz, while in the latter case, the dedicated actuator was used at an excitation frequency of 75 Hz, due to device upper freqeuncy limitations.

It is noteworthy that while different smartphones might vibrate at different freqeuncies and at different amplitude, the method presented overcomes this by being able to automatically select the freqeuncy or freqeuncies of interest using a peak detection algorithm. A potential challenging situation would happen if a strong noise signal would shadow the frequency of interest. Nevertheless, the method would still consider the three strongest occuring freqeuncies for the elastogram computation.

## IV. CONCLUSION

We showed in this paper that smartphones can be used as means of inducing harmonic excitations in tissue that produce large enough displacements which can be visualized and tracked using cross-correlation methods. A finite elemenent inverse-problem approach for elasticity reconstruction was used to successfully recover the Young's modulus of a CIRS quality assurance phantom with a circular inclusion. Similar results were obtained when compared to using a dedicated mechanical actuator. Due to the small vibration amplitude, the method is limited to surface organs, such as the breast or skin. Future studies will be performed to confirm the applicability of our low-cost ultrasound elastography method to diagnose breast or skin related pathologies in patients.

## ACKNOWLEDGMENT

This work was partly funded by the Swiss National Science Foundation. The author gratefully acknowledges the support from Prof. Orcun Goksel during his collaboration with the Computer-assisted Applications in Medicine group at ETH Zurich.